\documentclass[10pt,twocolumn,letterpaper]{article}

\usepackage{wacv}              

\usepackage{graphicx}
\usepackage{amsmath}
\usepackage{amssymb}
\usepackage{booktabs}
\usepackage{enumitem}
\usepackage{multirow}
\usepackage{diagbox}
\usepackage[figuresright]{rotating}
\usepackage{tabularx}
\usepackage{longtable}
\usepackage{makecell}
\usepackage{xcolor}

\usepackage[pagebackref,breaklinks,colorlinks]{hyperref}

\usepackage[capitalize]{cleveref}
\crefname{section}{Sec.}{Secs.}
\Crefname{section}{Section}{Sections}
\Crefname{table}{Table}{Tables}
\crefname{table}{Tab.}{Tabs.}

\def\wacvPaperID{289} 
\def\confName{WACV}
\def\confYear{2025}

\begin{document}

\title{Cross-domain Multi-modal Few-shot Object Detection via Rich Text}

\author{Zeyu Shangguan, Daniel Seita, and Mohammad Rostami\\
{\tt\small \{zshanggu,seita,rostamim\}@usc.edu}
\\
Department of Computer Science
\\
University of Southern California
}

\maketitle

\begin{abstract}
Cross-modal feature extraction and integration have led to steady performance improvements in few-shot learning tasks.
However, existing multi-modal object detection (MM-OD) methods degrade when facing significant domain shift and are sample insufficient.
We hypothesize that \textbf{rich text} information could more effectively help the model to build a knowledge relationship between the vision instance and its language description and can help mitigate domain shift.
Specifically, we study the \textbf{C}ross-\textbf{D}omain few-shot generalization of \textbf{MM-OD} (\textbf{CDMM-FSOD}) and propose a meta-learning based multi-modal few-shot object detection method that utilizes rich text semantic information as an auxiliary modality to achieve domain adaptation.
Our proposed novel neural network contains (i) a multi-modal feature aggregation module that aligns the vision and language support feature embeddings and (ii) a rich text semantic rectify module that utilizes bidirectional text feature generation to reinforce multi-modal feature alignment and thus to enhance the model's language understanding capability.
We evaluate our model on common standard cross-domain object detection datasets and demonstrate that our approach considerably outperforms existing FSOD methods. Our implementation is publicly available: \url{https://github.com/zshanggu/CDMM}
\end{abstract}

\section{Introduction}
\label{sec:intro}

In real-world industrial application, the majority of current deep learning-based object detection approaches encounter failures when detecting product defects~\cite{shangguan2024improved}, primarily due to the scarcity of relevant training data and the substantial domain gap between pre-trained datasets from scenarios common in daily life versus the specific domain of product defect data.
However, human experts are often equipped with comprehensive training manuals that provide detailed textual instructions and guidance, enabling them to discern object appearances and accurately annotate images. Consequently, in this paper, we propose a methodology that mimics this process, leveraging multi-modal rich textual information to augment the object detection task in scenarios where training data is scarce and out-of-domain.

\begin{figure}[t]
    \centering
    \includegraphics[width=\linewidth]{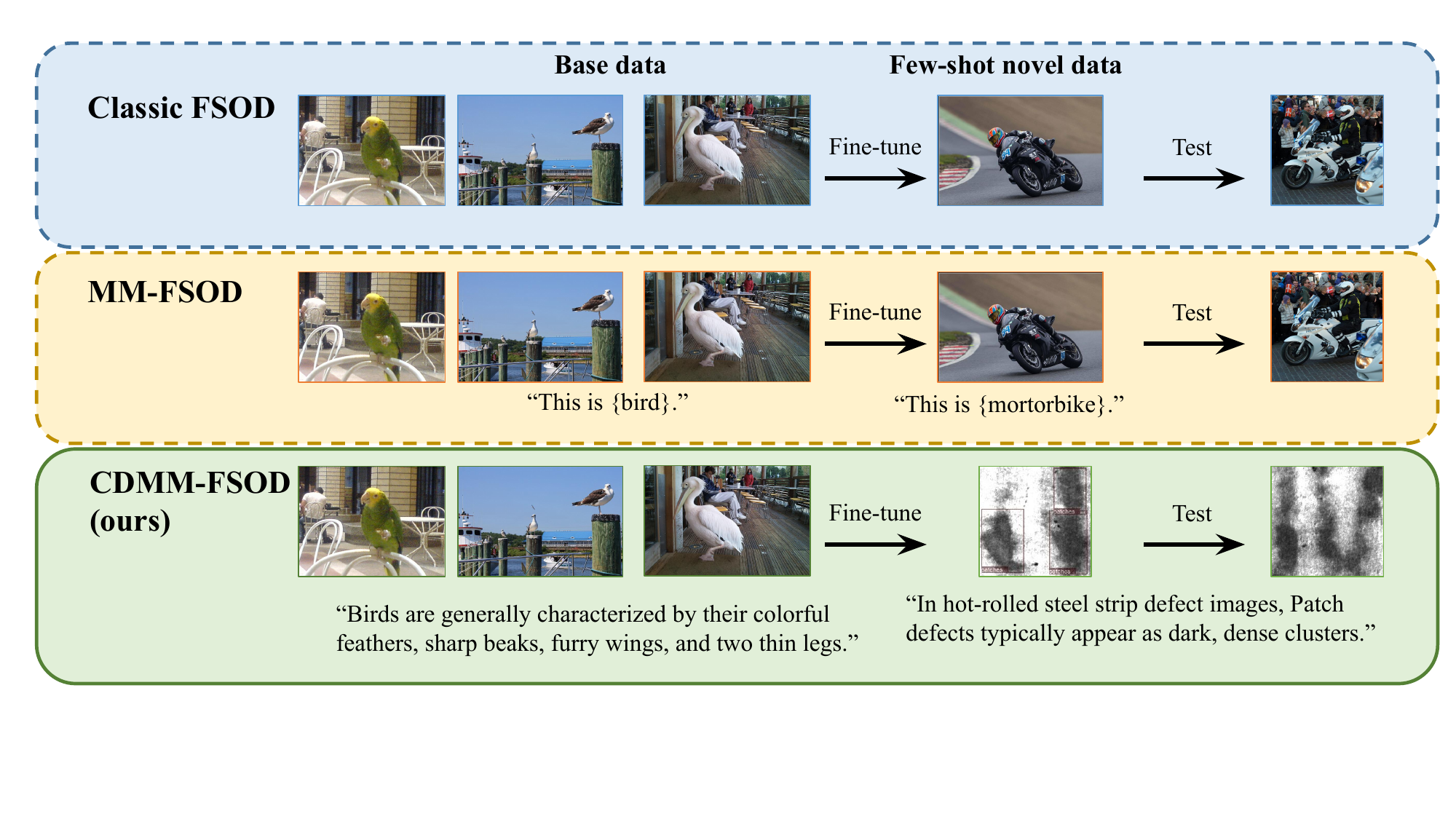}
    \vspace{-15pt}
    \caption{Different FSOD tasks. The classic FSOD task (top) involves only visual information. MM-FSOD (middle) introduces a language modality to provide extra information to improve FSOD performance. In contrast to these, our proposed CDMM-FSOD task (bottom) is tailored for cross-domain scenarios and extends MM-FSOD to use richer text. Above, we show a cross-domain example where a model might train on images and text of common data (such as birds) and needs to generalize to detection of less common, substantially different data (such as patch defects).}
    \label{fig:intro}
    \vspace{-15pt}
\end{figure}

\textbf{FSOD.} We study the general few-shot object detection (FSOD) task. The goal is to train a model that can generalize to detecting objects for which we only have a few labeled examples~\cite{Wang20TFA,Shangguan23HTRPN}.
A typical FSOD model is first pre-trained on a number of base classes for which we have abundant annotated data and then the model is fine-tuned on a set of novel classes with few annotated samples. 
In the classic setting, the base and the novel classes are generally within the same feature domain.
Multi-modal learning is a new approach for FSOD, and utilizes extra text information to enhance visual feature representations and provides potential zero-shot inference ability (see ~\cref{fig:intro}).
These methods encompass various techniques, including,  visual question-answering ~\cite{Han2024FMFSOD,cai2024clumo,zhang2024cross}, robotics \cite{rt22023arxiv,vision_touch_2019}, continual learning ~\cite{srinivasan2023i2i,cai2023task}, and medical image analysis ~\cite{Kline2022MultimodalML,moor2023med}.

\begin{figure}[t]
    \centering
    \includegraphics[width=\linewidth]{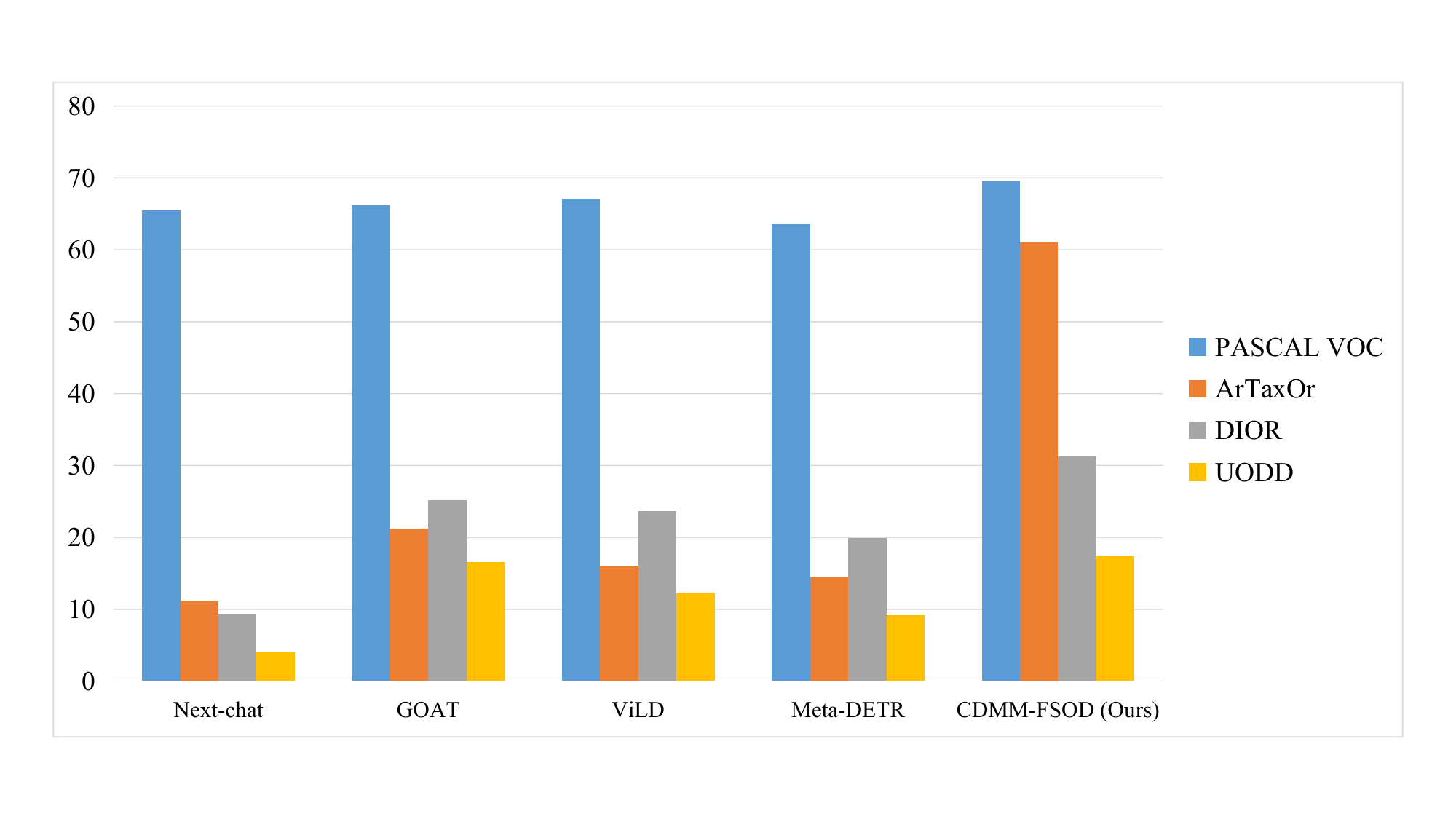}
    \vspace{-25pt}
    \caption{
    Performance results on 10-shot object detection on multiple cross-domain, few-shot datasets. We observe substantial cross-domain degradation for existing MM-OD Next-chat~\cite{zhang2023nextchat}, GOAT~\cite{Wang23GOAT}, and ViLD~\cite{gu2022openvocabulary}, as well as for a single-modal detection method, Meta-DETR~\cite{Zhang23MetaDETR}. In contrast, our proposed method has stronger performance on out-of-domain data. 
    }
    \label{fig:degradation}
    \vspace{-10pt}
\end{figure}

\textbf{Cross-domain problem in FSOD.} Despite the success of multi-modal object detection (MM-OD) methods on few-shot learning tasks, the source and the target data may have a substantial domain gap in many practical detection problems. Thus, the performance of MM-OD methods degrade in such cross-domain, data scarce scenarios. As shown in ~\cref{fig:degradation}, when we train and test object detection methods on data where the source and target domains differ, most MM-OD methods degrade significantly.
Therefore, our goal is to address the \textbf{cross-domain multi-modal few-shot object detection (CDMM-FSOD)} task, aiming to bridge the domain gap while transferring information from the source domain to the target domain to address this   challenge.

\textbf{Rich text.
We hypothesize that using rich text within a sentence can be more effective than relying solely on simple semantics, 
especially when images in the target domain require technical language to describe. 
However, for neural network models, how to utilize such rich text information for detection tasks remains unclear.
\emph{We aim to investigate whether and how rich text can address such cross-domain data gaps.} 
In~\cref{fig:intro}, we illustrate the difference of our proposed CDMM-FSOD compared to classic FSOD and MM-FSOD. Our rich text contains detailed technical descriptions of the training data, and our novel data (the patch defect in this example) has prominent domain shift. Given that this scenario is prevalent in industry, we believe our approach offers a way to tackle such practical challenges.}

\textbf{Our method.} We choose a meta-learning version of Detection Transformer (DETR)~\cite{Carion20DETR}, \ie Meta-DETR~\cite{Zhang23MetaDETR} as our baseline because of its Transformer-based structure and excellent modality adaption capacity.
Unlike the classic multi-modal setting, we define a rich text description for each training category which may contain technical terminology.
These sentences function as the linguistic support set for our meta-learning process, analogous to the role played by the image support set in terms of functionality.
We design a \textbf{meta-learning multi-modal aggregated feature module} to fuse the vision and the language embedding and map them to a class-agnostic feature space as the meta support feature.
Furthermore, to ensure the model obtains the detailed knowledge from the rich semantics, we propose a \textbf{rich semantic rectify module} to align the generated language embedding with the ground truth language embedding. 
In our experiments, we demonstrate consider performance improvement by using rich text. We further discuss and investigate the effect of text length on performance.

Our specific contributions include:
\vspace{-5pt}
\begin{itemize}[noitemsep,leftmargin=*]
    \item We propose a meta-learning multi-modal aggregated feature module that utilizes rich text semantics to alleviate the performance degradation due to domain gaps.
    \item We propose a rich semantic rectify module that learns to align the generated language embedding and the ground truth language embedding, enabling the model to better understand the text for knowledge transfer.
    \item Our experiments on cross-domain detection data demonstrate that our approach outperforms existing methods, suggesting that multi-modal learning with rich text is effective for detection of few-shot, out-of-domain data.
\end{itemize}

\section{Related Works}
\label{Related Works}

\subsection{Few-shot Object Detection}
\label{sec:FSOD}

The FSOD task is to train a model to accurately detect novel categories where only a few labeled samples are available. 
The dominant training paradigms of FSOD include methods based on fine-tuning and meta-learning. Fine-tuning methods often first pre-train a model on base data with sufficient training samples for each category, and then directly fine-tune the model on novel categories with few samples. Representative works include TFA~\cite{Wang20TFA}, FSCE~\cite{Sun21FSCE}, FSRC~\cite{Shangguan23FSRC}, DeFRCN~\cite{Qiao21DeFRCN}, \etc. 
In meta-learning methods, \ie learning to learn, instead of aiming at specific classes, a model is trained to learn a metric ability through tasks (\ie episodes), so that it can utilize a Siamese network~\cite{Koch2015SiameseNN} to achieve accurate classification by measuring the similarity between support and query features; thus, it is expected to learn class-agnostic knowledge~\cite{Chen21MetaBaseline}. 
Each training episode in meta-learning methods builds a support set and query set; an $n$-way $k$-shot support set consists of $n$ categories with $k$ training instances for each. 
Such work includes Meta-FRCN~\cite{han2022meta}, Meta-DETR~\cite{Zhang23MetaDETR}, etc. 

To avoid catastrophic forgetting~\cite{shmelkov2017incremental,feng2022overcoming,rostami2023cognitively} of the base categories, researchers tune FSOD models on samples of both base and novel classes.
However, while fine-tuning methods directly transfer the model on novel data in a class-specific manner, meta-learning methods tend to result in class-agnostic models, because they aim at learning a metric scheme between query features and support features. 
TFA~\cite{Wang20TFA} demonstrates that fine-tuning methods have potential to surpass meta-learning methods in detection performance. However, FCT~\cite{Han22FCT} shows that meta-learning is still a powerful solution for FSOD, and that such methods are better at rapidly adapting the model to new tasks, due to their task-level feature representation.

The traditional backbone networks for FSOD include Faster R-CNN~\cite{RenHGS15}, YOLO~\cite{redmon2016yolo}, and ViT~\cite{dosovitskiy2020vit}. More recently, Detection Transformer (DETR)~\cite{Carion20DETR, shangguan2023decoupled}, is gaining increasing popularity for FSOD due to its competitive performance compared to classical frameworks; furthermore, its transformer-based architecture facilitates integration with NLP tasks. In our work, we thus follow the meta-learning paradigm provided by Meta-DETR~\cite{Zhang23MetaDETR}.

\subsection{Prompt Learning and Multi-modal Few-shot Object Detection}
\label{sec:MM-FSOD}

Multi-modal object detection (MM-OD) extends the classic object detection task to incorporate sources of information beyond images. We consider language as the extra modality, so the task is   to train a model on image-text data. 
Most multi-modal object detection networks that benefit from prompt learning, utilize a pre-trained vision-language representation (\eg CLIP~\cite{Radford2021CLIP} is a standard choice. This strategy has been broadly adopted in open-vocabulary detection tasks (OVD), \ie the model could recognize novel classes with only a image-text pairs. Such a multi-modal representation ability greatly enhances the generalization of the base model, but usually utilizes relatively simple language, such as only using the category names or attributes~\cite{gu2022openvocabulary}. The development of Large Language Models has significantly ushered the development of multi-modal object detection. For example, Next-chat~\cite{zhang2023nextchat} proposes a Q\&A-based object detection-instance segmentation model. YOLO-World~\cite{cheng2024yoloworld} builds a visual-linguistic interactive model based on the idea of region-text contrastive.

Multi-modal few-shot object detection (MM-FSOD)  has not been broadly explored. Han \etal ~\cite{han2023multimodal} propose a framework that utilize soft prompt tokens to realize low-shot object detection. However, these methods   only use the simple image-level description or category name as the language label for the objects in an image. Such text information is insufficient when there exists a large domain gap when transferring knowledge from base classes to a novel dataset. Our work leverages this scheme to build a multi-modal network, using richer and more complex language. 

\subsection{Cross-domain Few-shot Object Detection}
Cross-domain few-shot object detection (CD-FSOD) aims at solving FSOD when the source and target feature domains are substantially different. MoFSOD~\cite{Lee22Rethink} proposes a multi-domain FSOD benchmark that consists of 10 datasets from different domains to simulate real-life situations. Their $k$-shot sampling is defined as $k$ images per class, and they demonstrate that effectively using the pre-training dataset plays a significant role in improving the model performance. CD-FSOD~\cite{Xiong2023CDFSOD} proposes a cross-domain FSOD benchmark with balanced data per category (\ie $k$ instances for each categories), making it better align with the general $K$-shot FSOD benchmark settings~\cite{Wang20TFA}. Furthermore, they propose a novel distillation-based baseline. Acrofod~\cite{Gao22CDFSOD} design an adaptive optimization strategy in selecting proper augmented data. We attempt to alleviate the cross-domain FSOD degradation issue with the help of multi-modal feature alignment, and we mainly follow the benchmark provided by~\cite{Gao22CDFSOD} to evaluate our model.

\section{Proposed Methods}
\label{sec:Proposed Methods}

\subsection{Preliminaries}
\label{sec:Preliminaries}


In the classical Few-Shot Object Detection (FSOD) task, there are a set of \emph{base} classes $C_B$ and \emph{novel} classes $C_N$. While each class in $C_B$ has abundant training data, each class in $C_N$ only has a small number of instances. We assume there is no overlap among classes in $C_B$ and $C_N$. The training data ${\mathcal{D}_{\rm FSOD} = \{(x,y)_i\}_{i=1}^{N_D}}$ consists of $N_D$ pairs of RGB images $x$ and ground-truth labels $y$. Here, each $y = \{(c_j, b_j) \mid j \in \{1,2,\ldots,N_{\rm obj} \}\}$ contains, for each of the $N_{\rm obj}$ object instances in image $x$, its class $c_j \in C_B \cup C_N$ and bounding box $b_j$. The aim is to train a model on $\mathcal{D}_{\rm FSOD}$ such that, at test time, it can effectively detect objects belonging to the novel classes~\cite{Kang19fsodrewei,Wang20TFA}.

The Multi-Modal Few-Shot Object Detection (MM-FSOD) task extends FSOD to incorporate text information. Formally, we have a dataset $\mathcal{D}_\text{MM-FSOD} = \{(x, \ell, y)_i\}_{i=1}^{N_D}$, where each data point now includes an additional set of category-specific text labels $\ell$ for each of the $N_{\rm obj}$ instances in image $x$. Formally, $\ell = \{l_j \mid j \in \{1, 2, \ldots, N_{\rm obj}\} \}$. Each text $l_j$ can take a variety of forms, and is typically either the category name or a template sentence that contains the name, such as: \texttt{This is a picture of <category>}. It is also possible to generate $\ell$ by using an external image-text model, such as CLIP~\cite{Radford2021CLIP}, to generate sentences from the image~\cite{han2023multimodal}. 
Finally, the Cross-Domain Multi-Modal Few-Shot Object Detection (CDMM-FSOD) task we study in this paper is an extension of MM-FSOD when the classes in $C_B$ and $C_N$ are substantially different. 

\subsection{Rich Text}
\label{sec:Rich Text}

We propose to manually define a fixed, rich text description for each category $c \in C_B \cup C_N$ to form the text $\ell$ for the training data for improved information content. 
We denote the set of text descriptions as $W^C=\{w^1, w^2, \ldots, w^C\}$, where $C = |C_B \cup C_N|$ is the number of categories.
Each text description contains multiple tokens with variable length of $M$, noted as $w^i = \{ w^i_1, w^i_2, \ldots, w^i_M\}$, where the first token $w^i_1 = \langle s \rangle$ is the start of the sentence, and $w^i_M = \langle /s \rangle$ is the end of the sentence.
To form the set $W^C$, we use two strategies: (i) we refer to the Wikipedia content of each category and define each category from the aspects of color, shape, attribute, material, etc. (ii) besides the aspects from the first strategy, we extend the sentence by adding the depiction of the image context while keep other text unchanged. For example, the original sentence ``Motorbikes have two wheels, a motor, and a sturdy frame'' will be extended as ``Motorbikes have two wheels, a motor, and a sturdy frame, usually ride by a person.'' In the revised sentence, we introduce a common visual relationship between two categories: motorbikes and persons. 

\begin{figure*}[t]
    \centering
    \includegraphics[width=0.7\linewidth]{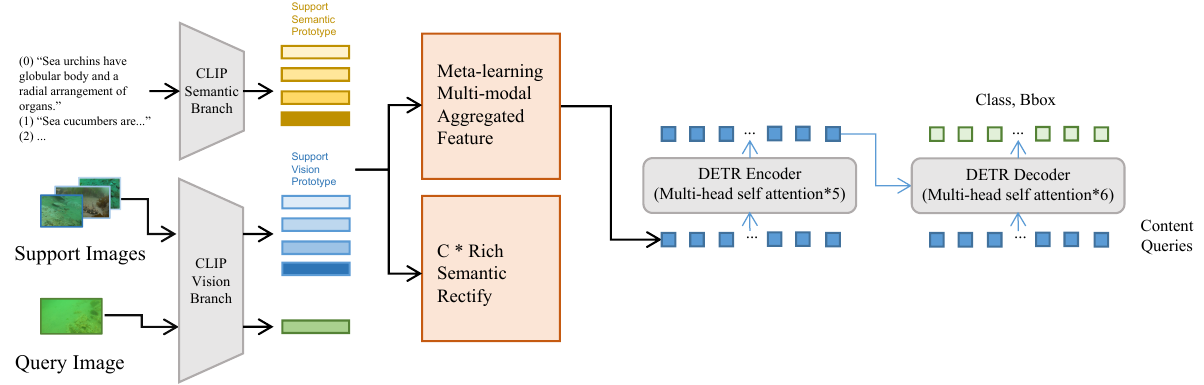}
    \vspace{-5pt}
\caption{The overall structure of our model. We indicate our proposed multi-modal feature aggregation module and rich text rectify module with the red blocks; see~\cref{fig:module} for more details about their structure. The multi-modal feature aggregation module is responsible for the cross-modality feature embedding mix. The rich text rectify module reinforces the model's cross-modality understanding. We design this end-to-end model that takes a set of support and query images, as well as a group of rich category text as input for training, and outputs the object detection results of the query images.
}
    \label{fig:pipeline}
\end{figure*}

\subsection{Proposed Architecture}
We use Meta-DETR~\cite{Zhang23MetaDETR} as our baseline network, which is built upon deformable DETR~\cite{zhu2021deformable}.
DETR is a transformer-based~\cite{vaswani2017attention} object detection model. It has a ResNet~\cite{ResNet2016} backbone to obtain the high-level image features; and then an encoder is applied to get the encoder memory embedding; followed by a decoder that could convert the memory embedding back to the object proposal.
For the backbone part, we follow Meta-DETR~\cite{Zhang23MetaDETR} and use a weight-shared self attention module to build the meta-learning scheme for the query and support images.
The classic DETR network includes an encoder and a decoder module, both consists of six multi-head attention layers. Meta-DETR build the meta-learning module only at the first multi-head attention layer of the encoder module. 

For the meta-learning module, the query and support images will be firstly sent to a shared ResNet feature extractor to get the basic image features. Then the query and support features will be passed through a multi-head attention module so that they are in the same feature space.
The support sample feature will then serve as the $K$ and $V$ matrices while the query image feature is $Q$ of a single-head self-attention module.
This single-head self-attention module is responsible for the class-specific feature matching among the support and query feature.
Meanwhile, in order to align with the class-agnostic learning task setting of meta-learning, this single-head self-attention module will be applied on a parallel branch to map the class-specific `support'-`query' relationship to a class-agnostic `task embedding'-`query' relationship.
Superficially, Meta-DETR designs an independent trainable class-agnostic feature prototype as $V'$ for the parallel single-head self-attention branch.
And the support and query sample feature will still serve as $K'$ and $Q'$ respectively.
In this way, the single-head self-attention module will achieve an important feature mapping so that the original class-specific support sample feature could be converted to class-agnostic prototype feature. 
During the test time, the model will only use the class-agnostic prototype as support features.
And then the aggregated class-agnostic prototype will be sent to the input of DETR to generate the final detection output.

\begin{figure*}[t]
  \centering
  \includegraphics[width=0.63\linewidth]{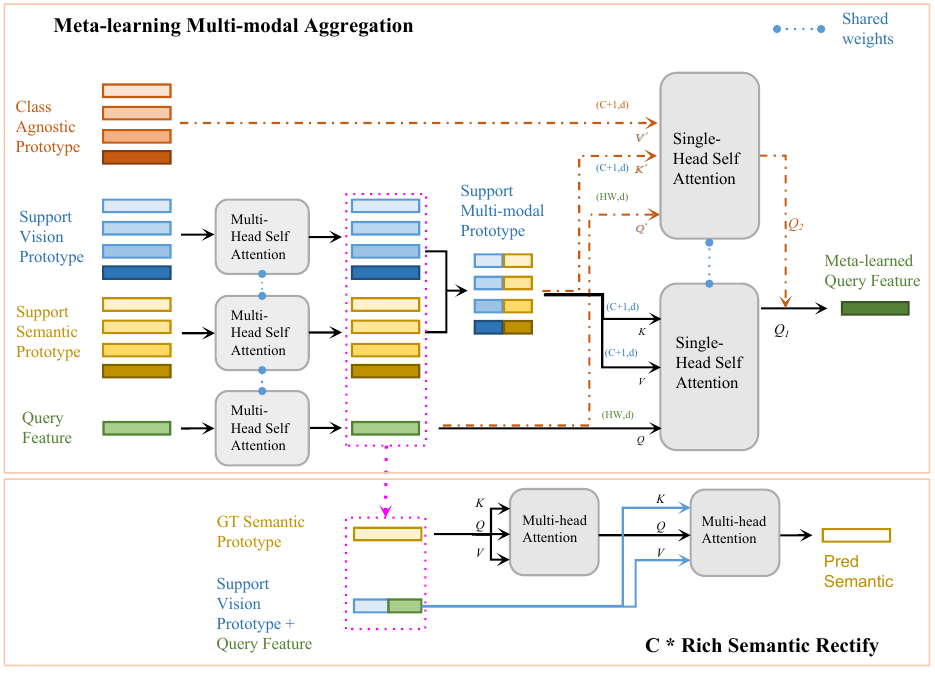}
    \vspace{-5pt}
  \caption{Details of our meta-learning multi-modal aggregation module (\emph{upper region}) and the rich semantic rectify module (\emph{lower region}). We use different colors for different feature branches. The rich semantic rectify module is only used during training, and not at test time. 
  }
  \label{fig:module}
    \vspace{-15pt}
\end{figure*}

The overall architecture of our proposed method is shown in~\cref{fig:pipeline}.
We add the rich text label as the auxiliary support feature.
Then we aggregate the image and language support features to train the single-head self-attention meta-learning module.
This module is the \textbf{meta-learning multi-modal feature aggregation module (MM Aggre.)}.
Besides, in order to ensure that the model properly learn the knowledge from the rich text information, we adopt a \textbf{rich text semantic rectify module (Rich Text Rect.)} to bidirectionally generate language feature, and then enforce it to align with the ground-truth language feature.

\begin{table*}[t]
\small
\centering
\resizebox{.9\textwidth}{!}{
\begin{tabular}{c|c|c|ccc|ccc|ccc}
\toprule
\multirow{2}{*}{\diagbox[height=20pt,innerrightsep=25pt]{\textbf{Method}}{\textbf{Shot}}} & \multirow{2}{*}{\textbf{Backbone}} & \multirow{2}{*}{\textbf{Modality}} & \multicolumn{3}{c|}{\textbf{ArTaxOr}} & \multicolumn{3}{c|}{\textbf{DIOR}} & \multicolumn{3}{c}{\textbf{UODD}} \\
			& & & 1 & 5 & 10 & 1 & 5 & 10 & 1 & 5 & 10 \\
\midrule

Meta-RCNN $\circ$ ~\cite{yan2019meta} & ResNet50 & Single & 2.8 & 8.5 & 14.0 & 7.8 & 17.7 & 20.6 & 3.6 & 8.8 & 11.2\\

TFA w/cos $\circ$~\cite{Wang20TFA} & ResNet50 & Single & 3.1 & 8.8 & 14.0 & 8.0 & 18.1 & 20.5 & 4.4 & 8.7 & 11.8 \\

FSCE $\circ$~\cite{Sun21FSCE} & ResNet50 & Single &  3.7 & 10.2 & 15.9 & 8.6 & 18.7 & 21.9 & 3.9 & 9.6 & 12.0 \\ 

DeFRCN $\circ$~\cite{Qiao21DeFRCN} & ResNet50 & Single & 3.6 & 9.9 & 15.5 & 9.3 & 18.9 & 22.9 & 4.5 & 9.9 & 12.1 \\ 

Distill-cdfsod $\circ$~\cite{Xiong2023CDFSOD} & ResNet50 & Single & 
5.1 & 12.5 & 18.1 & 10.5 & 19.1 & 26.5 & 5.9 & 12.2 & 14.5 \\

ViTDeT-FT$\dagger$ ~\cite{li2022exploring} & ViT-B/14 & Single & 5.9 & 20.9 & 23.4 & 12.9 & 23.3 & 29.4 & 4.0 & 11.1 & 15.6 \\

Detic-FT$\dagger$ ~\cite{zhou2022detecting} & ViT-L/14 & Multi & 3.2 & 8.7 & 12.0 & 4.1 & 12.1 & 15.4 & 4.2 & 10.4 & 14.4 \\
DE-ViT$\dagger$~\cite{zhang2023detect} & ViT-L/14 & Single & 0.4 & 10.1 & 9.2 & 2.7 & 7.8 & 8.4 & 1.5 & 3.1 & 3.1 \\
Meta-DETR$\dagger$~\cite{Zhang23MetaDETR} & DETR-R101 & Single & 6.5 & 11.3 & 14.5 & 11.1 & 19.4 & 19.9 & 5.8 & 9.5 & 9.2 \\
\midrule 
Next-Chat$\ddag$~\cite{zhang2023nextchat} & ViT-L/14 & Multi & 1.1 & 10.9 & 11.2 & 10.6 & 19.2 & 19.3 & 2.1 & 3.2 & 4.0 \\ 
GOAT$\ddag$~\cite{Wang23GOAT} & ResNet50 & Multi & 5.7 & 11.1 & 21.2 & 11.3 & 20.1 & 25.2 & 3.5 & 9.5 & 16.6 \\
ViLD$\ddag$~\cite{gu2022openvocabulary} & ResNet50 & Multi & 4.4 & 10.2 & 16.1 & 10.2 & 18.8 & 23.7 & 2.8 & 5.5 & 12.3 \\
\midrule
\textbf{Our Method} w/ self-built text & DETR-R101 & Multi & \textbf{14.8} & \textbf{48.1} & \textbf{61.0} & \textbf{13.7} & \textbf{26.3} & \textbf{31.3} & \textbf{5.5} & \textbf{11.9} & \textbf{17.4} \\
\textbf{Our Method} w/ LLM text & DETR-R101 & Multi & \textbf{15.1} & \textbf{48.7} & \textbf{61.4} & \textbf{14.3} & \textbf{26.9} & \textbf{31.4} & \textbf{6.9} & \textbf{12.5} & \textbf{17.5} \\
\bottomrule
\end{tabular}
}
    \vspace{-5pt}
\caption{Performance Results (mAP) on CD-FSOD benchmarks. The $\circ$ denotes results of general FSOD methods from Distill-cdfsod~\cite{Xiong2023CDFSOD}; $\dagger$ denotes that the methods are developed or the results are reported by Fu \etal ~\cite{fu2024cross}; $\ddag$ indicates the results of MM-FSOD   are reported by us. Highest scores are in bold font.}
\label{tab:main}
    \vspace{-10pt}
\end{table*}

\subsection{Multi-modal Feature Aggregation}
Multi-modal feature aggregation is the core module to fuse the multi-modal information, which is capable of aggregating multiple support categories.
We visualize its architecture in~\cref{fig:module}. 
We use the existing CLIP pre-trained model to extract the primary image feature of the support/query set and the primary semantic feature of the rich text.
Then the support image feature will be processed by a RoIAlign layer and an average pooling layer to get the instance-level category prototypes.
This is a common operation in object detection because the image feature could not represent the instance feature; there usually exists multiple foreground objects in one image.
Simultaneously, similar to prior object detection works, the background, as an extra hidden label, is also initialized as one of the category prototypes.
These features are then mapped to the same feature space through a shared multi-head self-attention module (\ie the first DETR encoder layer); our ablation study in~\cref{sec: ablation} verifies the importance of sharing weights. The rich text must be provided at inference time.

Then we concatenate and average the image and semantic features as the overall support feature (i.e. the support vision feature will be evenly concatenated with the corresponding language feature from the same category).
Then we use a single-head self-attention module to implement (i) the `support'-`query' feature mapping and (ii) the class-agnostic `task embedding'-`query' encoding matching.

For the the `support'-`query' feature mapping, the aggregated multi-modal feature will serve as key ($K$) and value ($V$) matrix for the single-head self-attention module.
Specifically, given $C$ support categories, for multi-modal support feature ($S \in \mathbb{R}^{(C+1) \times d}$) and query image feature ($Q \in \mathbb{R}^{HW \times d}$), the feature mapping coefficient will be:

\begin{equation}
    A = \text{softmax}\left(\frac{QS^T}{\sqrt{d}}\right)
\label{eq:feature mapping}
\end{equation}

In a query image, only certain foreground objects have matched feature with the support instances, while the remaining part (background) should be ignored. Therefore, we should only pay attention to certain area of the query feature map, and weaken those areas that do not match with any given support prototype. To this end, we apply a sigmoid function for support prototype $\sigma(S)$ as value matrix, and combine it with coefficient $A$ to build the attention mechanism.
Consequently, the output of the `support'-`query' feature mapping process would be:
\begin{equation}
    Q_1 = A\sigma(S)\odot Q,
\label{eq:output feature mapping}
\end{equation}
where $\odot$ represents the element-wise Hadamard product. The output $Q_1$ is a refined query feature map that keeps only the `foreground' features.

For the `task embedding'-`query' encoding matching, a set of class-agnostic task prototype ($T \in \mathbb{R}^{(C+1) \times d}$) is defined to replace the support class prototype. 
In other words, the class-specific support class prototype will be mapped to the pre-defined task prototype, so that the query feature will be matched with the task prototypes for the prediction process.
The task prototype $T$ would serve as the $V'$ matrix for the same single head self-attention module as the `support'-`query' feature mapping.
The initialization of $T$ is similar to sinusoidal position embedding in transformer.
The corresponding output would be:

\vspace{-10pt}
\begin{equation}
    Q_2 = AT
\label{eq:output task mapping}
\end{equation}

The overall output of the multi-modal aggregation module is the element-wise addition of $Q_1$ and $Q_2$.

\subsection{Rich Text Semantic Rectify}
The rich text semantic rectify module is depicted in ~\cref{fig:module}.
Our idea is to use a transformer encoder structure ~\cite{afham2021rich}, noted by $f_\theta$ to generate the category-level language sequence features of the fused support-query samples, and then the generated language feature is expected to be aligned with its ground truth feature.
Additionally, this language generation process is bidirectional, which means we generate the sentence in both forward and backward direction, so that we ensure the model has robust image-text understanding ability and encodes semantic knowledge. The Rich Semantic Rectify module is discarded at inference time.

For a given category $c$, its rich text tokens undergo refinement through the multi-head self-attention module that is shared with the vision branch to get its semantic prototype $l^c$, as shown in ~\cref{fig:module}.
Additionally, the support and query features are averaged to create a composite feature $p$, which serves as the $K$ and $V$ matrices for a multi-head self-attention module $A_3$, as shown in ~\cref{fig:module}.
In $A_3$, the refined text tokens act as the $Q$ matrix.
Subsequently, it generates the predicted text sequence $\hat W$ of length $M$.
This entire process is duplicated for both forward and backward prediction components, \ie the model will predict $\hat W$ from both the left-to-right and the right-to-left directions so that it could understand the rich text through its bidirectional context.
The loss function for this purpose would be:
{\small
\begin{equation}
\begin{split}
    \mathcal{L}_{rect} = \frac{1}{2} \left[\sum_{i=1}^{C}\sum_{j=2}^{M}-\mathrm{log} f_\theta(\hat w^i_j-w^i_j | p) +
    \right. \\ \left.
    \sum_{i=1}^{C}\sum_{j=1}^{M-1}-\mathrm{log} f_\theta(\hat w^i_j-w^i_j | p) \right]
\end{split}
\label{eq:rectify loss}
\end{equation}
}

\section{Experiments}
\subsection{Implementation Details}
\label{sec:Implementation}

We conduct our experiments on hardware consisting of 8 3090 GPUs for parallel training and testing.
We adopt hyper-parameter values from DETR~\cite{Carion20DETR} by setting the initial learning rate as 0.001, and using a batch size of 4 for 1-shot scenarios and 1 for 5-shot and 10-shot scenarios. The DETR encoder and decoder both have 6 self-attention layers (the first layer of the encoder is embedded in our meta-learning multi-modal aggregation module). 
 Following the literature, we rely on Afham \etal \cite{afham2021rich} for the architecture of the image-text transformer decoder.

\begin{table*}[t]
	\centering
	\resizebox{1\textwidth}{!}{
		\begin{tabular}{l|l|lllll|lllll|lllll}
			\toprule
			\multirow{2}{*}{\diagbox[height=20pt,innerrightsep=25pt]{Method}{Shot}} & \multirow{2}{*}{Backbone} & \multicolumn{5}{c|}{Split1} & \multicolumn{5}{c|}{Split2} & \multicolumn{5}{c}{Split3} \\
			& & 1 & 2 & 3 & 5 & 10 & 1 & 2 & 3 & 5 & 10 & 1 & 2 & 3 & 5 & 10\\
			\midrule
    TIP \cite{Li21TIP} & FRCN-R101 & 27.7 & 36.5 & 43.3 & 50.2 & 59.6 & 22.7 & 30.1 & 33.8 & 40.9 & 46.9 & 21.7 & 30.6 & 38.1 & 44.5 & 50.9 \\
			CME \cite{Li21CME} & FRCN-R101 & 41.5 & 47.5 & 50.4 & 58.2 & 60.9 & 27.2 & 30.2 & 41.4 & 42.5 & 46.8 & 34.3 & 39.6 & 45.1 & 48.3 & 51.5 \\
			DC-Net \cite{Hu21DCNet} & FRCN-R101 & 33.9 & 37.4 & 43.7 & 51.1 & 59.6 & 23.2 & 24.8 & 30.6 & 36.7 & 46.6 & 32.3 & 34.9 & 39.7 & 42.6 & 50.7 \\
           CGDP \cite{Li21CGDP} & FRCN-R2101 & 40.7 & 45.1 & 46.5 & 57.4 & 62.4 & 27.3 & 31.4 & 40.8 & 42.7 & 46.3 & 31.2 & 36.4 & 43.7 & 50.1 & 55.6 \\
        Meta Faster R-CNN \cite{han2022meta} & FRCN-R101 & 40.2 & 30.5 & 33.3 & 42.3 & 46.9 & 26.8 & 32.0 & 39.0 & 37.7 & 37.4 & 34.0 & 32.5 & 34.4 & 42.7 & 44.3 \\
           FCT \cite{Han22FCT} & PVTv2-B2-Li & 49.9 & \textbf{57.1} & 57.9 & 63.2 & 67.1 & 27.6 & 34.5 & 43.7 & 49.2 & 51.2 & 39.5 & \textbf{54.7} & 52.3 & 57.0 & 58.7 \\
           Meta-DETR \cite{Zhang23MetaDETR} & DETR-R101 & 40.6 & 51.4 & 58.0 & 59.2 & 63.6 & 37.0 & 36.6 & 43.7 & 49.1 & 54.6 & 41.6 & 45.9 & 52.7 & 58.9 & 60.6 \\
           FM-FSOD \cite{Han2024FMFSOD} & ViT-S & 41.6 & 49.0 & 55.8 & 61.2 & 67.7 & 34.7 & 37.6 & 47.6 & 52.5 & \textbf{58.7} & 39.5 & 47.8 & 54.4 & 57.8 & 62.6 \\
			MM-FSOD \cite{han2023multimodal} & \multirow{1}{*}{ViT} & 42.5 & 41.2 & 41.6 & 48.0 & 53.4 & 30.5 & 34.0 & 39.3 & 36.8 & 37.6 & 39.9 & 37.0 & 38.2 & 42.5 & 45.6 \\
			\midrule
			\textbf{Our Method} & DETR-R101 & \textbf{45.1} & 54.9 & \textbf{62.2} & \textbf{65.1} & \textbf{69.6} & \textbf{41.7} & \textbf{42.1} & \textbf{47.7} & \textbf{54.3} & 57.3 & \textbf{49.2} & 52.2 & \textbf{56.9} & \textbf{63.1} & \textbf{64.0} \\
			\bottomrule
		\end{tabular}
  }
\vspace{-5pt}
\caption{Performance results on PASCAL VOC dataset w/ self-built rich text descriptions. Highest scores are bolded.}
\label{tab:voc}
    \vspace{-5pt}
\end{table*}

\subsection{Datasets}
\label{sec:datasets}

Following the standard evaluation on the CD-FSOD benchmark provided by Xiong \etal ~\cite{Xiong2023CDFSOD}, we use ArTaxOr~\cite{Drange2020arthropod}, DIOR~\cite{LI2020DIOR} and UODD~\cite{Jiang2021UODD} as our target domain few-shot datasets.
ArTaxOr contains 7 arthropods categories, DIOR consists of 20 satellite images categories, and UODD has 3 underwater creature categories. 

\begin{table*}[ht]
    \centering
	\resizebox{1\textwidth}{!}{
    \begin{tabularx}{\textwidth}{l|X|c}
        \toprule
        Type of text & Example & UODD \\
        \midrule
        None & \eg None &  9.2 \\
        \midrule
        Category name & \eg Sea cucumbers &  9.9 \\
        \midrule
        Rich text & \eg Sea cucumbers have sausage-shape, usually resemble caterpillars; their mouth is surrounded by tentacles & 17.4 \\
        \midrule
        Extended rich text & \eg Sea cucumbers have sausage-shape, usually resemble caterpillars; their mouth is surrounded by tentacles; usually seen together with sea urchins. & 16.9 \\
        \midrule
        LLM-Rich text & \eg Sea cucumbers are marine invertebrates known for their elongated, leathery bodies, which are typically covered in spines or tentacles and lack a distinct head or tail & 17.5 \\
        \bottomrule
    \end{tabularx}
    }
    \vspace{-5pt}
    \caption{The effect of different language modalities.}
    \label{tab:text length}
    \vspace{-10pt}
\end{table*}

To measure the generalization of our method, we evaluate on the general FSOD benchmark of PASCAL VOC. Following the pioneering evaluation pipeline for FSOD~\cite{Wang20TFA}, 15 of the categories of PASCAL VOC are selected as base classes for pre-training, and the remaining 5 categories are used as the novel classes for few-shot fine-tuning. 


\subsection{Results on CD-FSOD Datasets}
Our experimental results on CD-FSOD datasets are presented in ~\cref{tab:main}. We follow the experimental setting from Xiong et~al.~\cite{Xiong2023CDFSOD}; our model is pre-trained on COCO and fine-tuned on the three out-of-domain datasets. We report the average precision (mAP) for evaluation. Since there is no existing work on CDMM-FSOD, we implement representative state of the art FSOD and MMOD works on the CD-FSOD benchmark and compare our method with them. The values reported by Distill-cdfsod~\cite{Xiong2023CDFSOD} and Fu \etal ~\cite{fu2024cross} are directly from the papers; while the values of other methods (\ie Meta-DETR, Next-Chat, GOAT and ViLD) are generated by us. We run the official code of these methods on the CD-FSOD benchmark. The experimental results indicate that our method outperforms the baselines. It is worth noting that we achieve significant improvement on the ArTaxOr dataset. Specifically, with LLM text, we reach 15.1, 48.7, 61.4 for 1,5,10-shot scenarios compared to the highest baseline values of 6.5, 20.9, 23.4, respectively. The reason is that the 7 categories of ArTaxOr have less domain shift compared to the COCO pre-training dataset in terms of the more clearer object boundary compared to other two cross-domain benchmarks~\cite{fu2024cross}.

As introduced in ~\cref{sec:Rich Text}, we build rich text manually or by using an LLM~\cite{Sun2021ERNIE3L}. As indicated in ~\cref{tab:main}, either way could generate significant improvement on the out-of-domain benchmarks. It is observed that LLM-generated rich text brings better performance on these three benchmarks, because the generated text contains more comprehensive sentences. However, this is not a universal situation, because sometimes the LLM failed to generate distinct descriptions for the target categories when their names are abstract and technical. 

\subsection{Results on Standard FSOD Datasets}
As an auxiliary analysis and to further evaluate the generalization of our method, we test our model on a common FSOD benchmark, \ie few-shot PASCAL VOC, in ~\cref{tab:voc}. Following TFA~\cite{Wang20TFA}, as described in ~\cref{sec:datasets},  there are three fixed groups of base-novel category splits, namely split 1, 2, and 3 to avoid sampling bias. For each data split, the $n$-shot ($n \in \{1,2,3,5,10\}$) sampling for the novel categories are also pre-defined. The model is evaluated on these three respective dataset splits. In addition, as described in ~\cref{sec:Preliminaries}, the base categories are also trained during fine-tuning to avoid forgetting, but have different sampling strategies: (i) in works such as TFA~\cite{Wang20TFA}, the base categories also follow the restricted $n$-shot sampling (\ie $n$ instances instead of $n$ images for each categories) as the novel categories; (ii) while in works such as Meta-DETR~\cite{Zhang23MetaDETR}, the base categories are not few-shot. Both strategies are acceptable for the FSOD task; we adapt to the later one. As shown in ~\cref{tab:voc}, our method outperforms previous works.

\subsection{Ablation Study}
\label{sec: ablation}

\textbf{Rich text length.} We first verify that the length of the rich text has a distinct impact on the model's performance. We conduct experiments on the UODD dataset using 4 cases of text: (i) the raw category name, (ii) our manually-built rich text referencing Wikipedia, (iii) an extended version of (ii) by adding image context, (iv) LLM-generated rich text. Cases (ii) to (iv) all result in long sentences that have rich semantics, and thus are generally beneficial for the model performance. 
See ~\cref{tab:text length} for our examples and quantitative results. 
However, not all extensions on the sentence length are effective. For example, in case (iii), our extended rich text for sea cucumbers has a slightly negative impact over the non-extended version, decreasing performance from 17.4 to 16.9. This may be because the extra image context with an distinct name of another category in UODD (\ie sea urchin) has confused the model.

\textbf{Module analysis.} In addition, we analyze the effectiveness of our proposed multi-modal aggregated feature module and the rich semantic rectify module. As seen in ~\cref{tab:modules}, comparing to the single-modal baseline, our method can achieve significant improvement through using rich text multi-modal information.  Our rich text rectify module also leads to an additional improvement.

\textbf{Which modality dominates the result.} We conducted this ablation study in ~\cref{tab:modality}. When only the image modality is utilized, the model corresponds to the baseline configuration. When only the language modality is incorporated, we leverage the CLIP language branch to extract prototype features. Our findings indicate that relying solely on the language modality to generate support prototypes leads to a performance decrease. The reason is that the language modality is prepared for each category, instead of for each image, and thus cannot provide a task-level support feature that is typically seen in general meta-learning settings. However, in ~\cref{tab:modules}, with the help of the language prototype, we achieve great improvement, as CLIP provides strong vision-language modality feature alignment.

\textbf{Shared vs decoupled attention.} In our MM Aggre. module, we use shared attention module. Alternatively, we could use decoupled attention module, hence, we compare their performance. For decoupled attention, we separate the multi-head self-attention module for the support vision and semantic prototypes. As shown in ~\cref{tab:decoupled attention}, the model's performance deteriorated with decoupled attention modules. As mentioned in ~\cref{sec:Rich Text}, the language prototypes are designed at the category level, not the image level. Thus, for the task-level support set, a separate attention module tends to diminish the gain from the language modality.

\begin{table}[t]
    \centering
	\resizebox{\linewidth}{!}{
    \begin{tabular}{l|c|c|c}
        \toprule
        Module & ArTaxOr & DIOR & UODD \\
        \midrule
        Meta-DETR & 14.5 & 19.9 & 9.2 \\
        Meta-DETR + MM Aggre. & 59.8 & 30.1 & 15.7 \\
        Meta-DETR + MM Aggre. + Rich Text Rect. & 61.4 & 31.4 & 17.5 \\
        \bottomrule
    \end{tabular}
    }
    \vspace{-5pt}
    \caption{The effect of different modules.}
    \label{tab:modules}
    \vspace{-10pt}
\end{table}

\begin{table}
    \centering
    \resizebox{.9\linewidth}{!}{
    \begin{tabular}{c|c|c|c}
        \toprule
        Method & ArTaxOr & DIOR & UODD \\
        \midrule
        Image modality only & 14.5 & 19.9 & 9.2 \\
        Language modality only & 11.3 & 7.7 & 5.1 \\
        \bottomrule
    \end{tabular}
    }
    \vspace{-5pt}
    \caption{\small Ablation on single modules.}
    \label{tab:modality}
    \vspace{-10pt}
\end{table}

\begin{table}
    \centering
    \resizebox{.9\linewidth}{!}{
    \begin{tabular}{c|c|c|c}
        \toprule
        Method & ArTaxOr & DIOR & UODD \\
        \midrule
        Shared attention & 61.4 & 31.4 & 17.5 \\
        Decoupled attention & 46.1 & 17.7 & 12.4 \\
        \bottomrule
    \end{tabular}
    }
    \vspace{-5pt}
    \caption{Ablation on shared vs decoupled attention.}
    \label{tab:decoupled attention}
    \vspace{-10pt}
\end{table}

\begin{figure}[t]
    \centering
    \includegraphics[width=.95\linewidth]{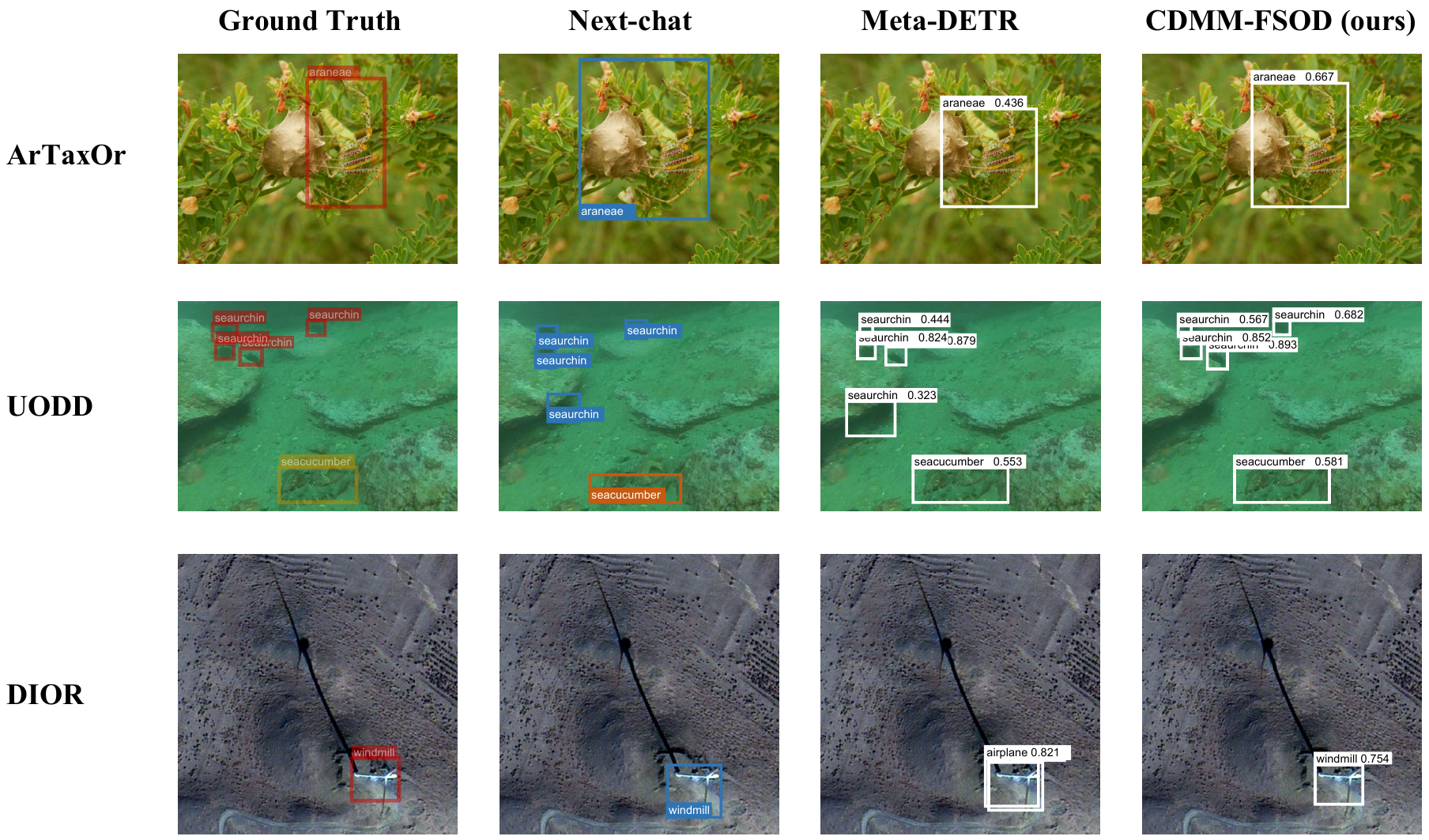}
    \vspace{-10pt}
\caption{Representative visualizations of detection results on three benchmark datasets. We compare our method with a multi-modal object detection model (Next-chat) and a few-shot object detection model (Meta-DETR). Our proposed model obtains more accurate bounding boxes and improved detection confidence.}
    \label{fig:visualization}
    \vspace{-15pt}
\end{figure}

\subsection{Visualization}
\label{sec:visualization}

In ~\cref{fig:visualization}, we present representative qualitative detection results on the three CD-FSOD benchmarks. In these examples, the objective is to detect araneae (ArTaxOr), sea urchins and sea cucumbers (UODD), and windmills (DIOR). 
The inference uses a confidence threshold setting of 0.3 for all methods. 
Compared to baseline methods, we show qualitative improvements in a more accurate bounding box (first row), fewer false positives (second row), and improved detection confidence (third row). 

\subsection{Discussion}
We observe that it is desirable when the rich text description of the target objects are available, but the specific context of the rich text plays a critical role. Consequently, in practice, the user must carefully evaluate the rich text and substantially tune it for greater improvement. This might become a burden in some cases when (i) the user has a rich text, but it contains many descriptions that are irrelevant to the visual information, or (ii) the user does not have rich text and thus need to build one from scratch. For the first case, a general suggestion for practical usage is to keep the information most related to the appearance of the target objects. For the second case, a user could extend and improve upon our LLM-generated text approach. In future work, for example, we plan to train a discriminator to select the best rich text from multiple candidates.

\section{Conclusion}
We introduce a   cross-domain multi-modal few-shot object detection network. We verify that using rich text information is a promising strategy for robust detection in out-of-domain, few-shot scenarios. Our experiments indicate that the design of rich text is a key impact factor for the  performance, and that a precise description of the appearance of the target objects is especially effective for few-shot learning. Experimental results demonstrate improved precision over current state-of-the-art methods on multiple cross-domain object detection datasets. We hope that this paper inspires future work to explore using multi-modality for bridging domain gaps in other computer vision tasks.


{\small
\bibliographystyle{ieee_fullname}
\bibliography{egbib}
}

\end{document}